\documentclass[conference]{IEEEtran}
\IEEEoverridecommandlockouts
\usepackage{cite}
\usepackage{amsmath,amssymb,amsfonts}
\usepackage{algorithmic}
\usepackage{graphicx}
\usepackage{textcomp}
\usepackage{xcolor}
\usepackage{graphicx}
\usepackage{subcaption}
\usepackage{booktabs}
\usepackage[utf8]{inputenc}

\def\BibTeX{{\rm B\kern-.05em{\sc i\kern-.025em b}\kern-.08em
    T\kern-.1667em\lower.7ex\hbox{E}\kern-.125emX}}
\begin{document}

\title{Determining the Optimal Number of Clusters for Time Series Datasets with Symbolic Pattern Forest\\
{\footnotesize \textsuperscript{}}
\thanks{}
}

\author{\IEEEauthorblockN{Md Nishat Raihan}
\IEEEauthorblockA{\textit{Department of Computer Science} \\
\textit{George Mason University}\\
Fairfax, Virginia \\
mraihan2@gmu.edu}
}

\maketitle

\begin{abstract}
Clustering algorithms are among the most widely used data mining methods due to their exploratory power and being an initial preprocessing step that paves the way for other techniques. But the problem of calculating the optimal number of clusters (say k) is one of the significant challenges for such methods. The most widely used clustering algorithms like k-means and k-shape in time series data mining also need the ground truth for the number of clusters that need to be generated. In this work, we extended the Symbolic Pattern Forest algorithm, another time series clustering algorithm, to determine the optimal number of clusters for the time series datasets. We used SPF to generate the clusters from the datasets and chose the optimal number of clusters based on the Silhouette Coefficient, a metric used to calculate the goodness of a clustering technique. Silhouette was calculated on both the bag of word vectors and the tf-idf vectors generated from the SAX words of each time series. We tested our approach on the UCR archive datasets, and our experimental results so far showed significant improvement over the baseline. 
\end{abstract}


\section{Introduction}
Clustering is a solution for classifying massive data when there is no prior knowledge about categories. With new ideas like big data and their expansive applications in current years, research has grown on unsupervised solutions like clustering algorithms to pull knowledge from this massive amount of data. Time-series clustering has been used in myriad scientific areas to find patterns that enable data analysts to gather useful information from complicated and gigantic datasets. In the case of massive datasets, using supervised solutions is almost incomprehensible, while clustering algorithms can crack this problem using unsupervised approaches. The time series clustering problem can be formulated as follows - given a set of unlabeled time series instances, the objective is to place them into separate, homogeneous groups.

\par

\begin{figure}[t]
  \centering
  \includegraphics[width=\linewidth]{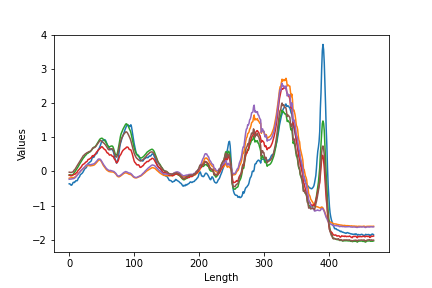}
  \caption{Beef Dataset from UCR Archive\cite{b5}}
\end{figure}

With the progress of technologies, for example, more delicate, smaller, and more affordable sensors widely embedded in different devices and machines, the quantity of time series data becomes massive, and the scope is changing rapidly. This demands the data mining algorithms to have low time complexity. Although there has been much work on time series clustering, little work is on providing a linear time solution with reasonable performance. Existing super-linear time complexity methods may not apply when the dataset is large or when real-time analytics are required. In this work, we extended the Symbolic Pattern Forest (SPF) algorithm proposed by Xiaosheng et al. \cite{b1}, which has linear time complexity. SPF checks if some randomly selected symbolic patterns exist in the time series to partition the data instances. This partition process is executed multiple times, and the partitions are combined by an ensemble process to generate the final partition. Similar to other widely used clustering algorithms in time series data mining like K-means, KSC \cite{b2} and K-shape \cite{b3}, Symbolic Pattern Forest (SPF) also requires the ground truth value for the actual number of clusters to work. Hence, we used SPF to generate multiple number of clusters from the datasets and then predicted the optimal number of cluster (optimal K). 

\par
We used the Silhouette Coefficient as the clustering metric in this work. The silhouette score for each value of K determines how good that number of cluster actually is for the particular dataset. It can range from -1 to 1, the higher being the better. Initially, we calculated the silhouette scores for different number of clusters on raw time series data. These baseline results were very inconsistent with unsatisfactory results. 

\par
In our approach, we used the Symbolic aggregate approximation (SAX) algorithm by Lin et al. \cite{b4} to transform each time series sub-sequences to a symbolic pattern. SAX converts the time series sub-sequences into words and with all the SAX words we generated the Bag of Words vectors from them. Each Bag of Words vector are of the same length of the total number of words found in the whole dataset and each value in the vector represents the total number of times the corresponding word appeared in that specific time series. We then calculated the Silhouette score for all the clusters using these vectors and picked K with the highest silhouette score. The results were consistent and had significant improvement over the baseline results.

\par
In addition to the Bag of Word vectors, we also generated the TF-IDF vectors based on the SAX words. TF-IDF vectors are almost similar to Bag of Word vectors, but they also serve two more purposes - they give more weights to the words that are comparatively rare and they give more importance to the words that are more frequent. These vectors were also used to calculate the Silhouette score and based on the scores, the optimal value for K was chosen. The results were a bit better than the ones with Bag of Word vectors, which also means a significant performance gain over the baseline.

\par
In this work, the primary contributions are -

\begin{itemize}
    \item extending the Symbolic Pattern Forest \cite{b1} algorithm to predict the optimal number of clusters when the ground truth for the actual number of clusters is not known.
    \item showing with experimental data that, even though Silhouette Coefficient is widely used as a clustering metric to determine the number of clusters, it is not an efficient metric to use on raw time series datasets. Rather silhouette on our methodology where we used the SAX words to generate Bag of Words vectors and TF-IDF vectors generates much better results than the baselines.
\end{itemize}

\par
The rest of the paper is organised as follows, section II provides the background and related works, section III describes the algorithms we used (SPF\cite{b1} and SAX\cite{b4}) briefly, section IV describes our methodology, section V describes the experimental results and finally section VI concludes the paper.

\section{Literature Review}

\subsection{Definitions and Notations}

This subsection provides the definitions and notations to precisely describe the problem under investigation and to present the proposed method.

\begin{itemize}
    \item \textbf{Definition 1:} A time series T is a ordered sequence of real-value data points [t\textsubscript{1}, t\textsubscript{2}, . . . , t\textsubscript{m}], where m is the length of the time series.
    \item \textbf{Definition 2:} A subsequence S of time series T is a sequence of contiguous values taken from T : S = [t\textsubscript{i}, t\textsubscript{i+1}, . . . , t\textsubscript{i+l-1}], where l is the length of the subsequence, $1 \leq i \leq m - l + 1 $ and $1 \leq l \leq m$. All subsequences of a certain length from a time series can be extracted using a sliding window of the same length from the first data point to the (m - l + 1) th point.
    \item \textbf{Definition 3:} Given a set of time series ${T\textsubscript{i}}\textsuperscript{n}\textsubscript{i=1} $, where n is the number of time series instances, time series partitional clustering assigns a group relationship c\textsubscript{i} for each T\textsubscript{i}, with c\textsubscript{i} = r\textsubscript{j}, $j \in {1, 2, . . . , k}$. r\textsubscript{j} is a group value and k is the number of clusters. Usually we have $k \ll m$ and $k \ll n$. For presentation simplicity, we assume all the time series in the dataset have the same length m. The proposed algorithm in the paper can also work on datasets with varying-length time series.
\end{itemize}

\subsection{Related Works}

K-means clustering algorithm \cite{b6} is one of the most widely used algorithms in clustering. It first starts by choosing k samples (objects) as the initial centroids. After that, each sample in the dataset is assigned to the nearest centroids based on a particular proximity measure. The most frequently used are Manhattan, Euclidean, and Cosine distances. Once the clusters are formed, the centroids are updated. The algorithm iteratively performs the assignment and update steps until a convergence criterion is met. 

\par

In the case of the standard k-means clustering algorithm, as the distance metric, Euclidean Distance (ED) \cite{b7} is used, and arithmetic means are adopted to calculate the centroids. However, it is not  uncommon for real-world time-series data to contain phase shifts, distortions, warpings, and amplitude changes. The simple Euclidean Distance might not be able to handle these situations. Therefore, many time series distance measurement techniques are suggested \cite{b8}, and Dynamic Time Warping (DTW) \cite{b9} being the most popular one, which can align the data points from the two time series under comparison to find the optimal comparison.

\par

However, in case of k-means and other algorithms that are based on k-means, they initially require the value for K - the total number of clusters, as they are going to generate k number of centroids and perform the clustering around those centroids. Algorithms like, K-Spectral Centroid (KSC) \cite{b10} that suggests a distance measure which looks for the optimal alignment and scaling for comparing two time series.  Also, K-shape \cite{b11} is one of the state-of-the-art time series clustering algorithms and it is also based on k-means algorithm. It suggests a new distance measure which is called Shape Based Distance (SBD), that is based on the time series cross-correlation. In this case, the centroids are generated by optimizing the within-cluster squared normalized crosscorrelation between the centroids and the time series instances. But both KSC\cite{b10}  and K-shape\cite{b11} are based on K-means and need the number of centroids to begin with. In another work by Zakaria et al. \cite{b12}, the authors suggest to itemize all the subsequences in the time series dataset to select a subset of subsequences called U-shapelets that can best separate the data. The distances between the time series and these subsequences are computed and regarded as new feature values. Finally, they used k-means, meaning the need for the number of clusters.

\par

There are few methods to determine the optimal number of clusters for k-means and the algorithms that are based on it. The oldest one among them them is the Elbow Method \cite{b13} where the idea is to start with K=2, and keep increasing it in each step by 1, calculating the clusters and the total cost that comes with the training. At some value for K the cost drops drastically, and after that it reaches a plateau when increased further and this is the chosen value for K. There also are other methods like  Calinski Harabasz Index \cite{b14}, Davies Bouldin Index \cite{b15} and Silhouette Coefficient. The Silhouette Coefficient is widely used as a clustering metric in recent years for a wide variety of datasets, in the works of Aranganayagi et al. \cite{b16}, Dinh et al. \cite{b17}, Shahapure et al. \cite{b18} etc.

\par

To the best of our knowledge, Silhouette coefficient was not used to determine the optimal number of clusters for any time series datasets in any of the works. In our work, initially, we experimented with Silhouette Coefficient to choose the optimal number of clusters for the time series data and the results on the raw time series were very inconsistent. So, in our actual approach we converted the time series to their corresponding Bag of Word Vectors and TF-IDF vectors and tested Silhouette on them, which generated a lot better results. We extended the work of Xiaosheng et al. \cite{b1} who obtained better results in time series clustering than most of the traditional algorithms like K-Means \cite{b6} and K-shape\cite{b3} and also has a linear time complexity but requires the ground truth value for K like other algorithms. In this work, we used the Silhouette scores on the vectors that we generated from the original time series and used that to determine the optimal number of cluster by using the Symbolic Pattern Forest \cite{b1} algorithm. Our experimental results on the datasets from the UCR archive \cite{b5} shows a significant improvement in results compared to the baseline.

\section{Symbolic Aggregate Approximation and Symbolic Pattern Forest}

\subsection{Symbolic Aggregate Approximation} \label{SAX}

Since our method uses Symbolic Aggregate approXimation (SAX) \cite{b4} to transform a time series subsequence to a symbolic pattern, we briefly describe this technique. Figure 2 shows an example of transforming a subsequence to a symbolic pattern (SAX word). The subsequence is z-normalized and divided into $\omega$ segments ($\omega$ is 2 in this example). 

\begin{figure}[httb!]
  \centering
  \includegraphics[width=\linewidth]{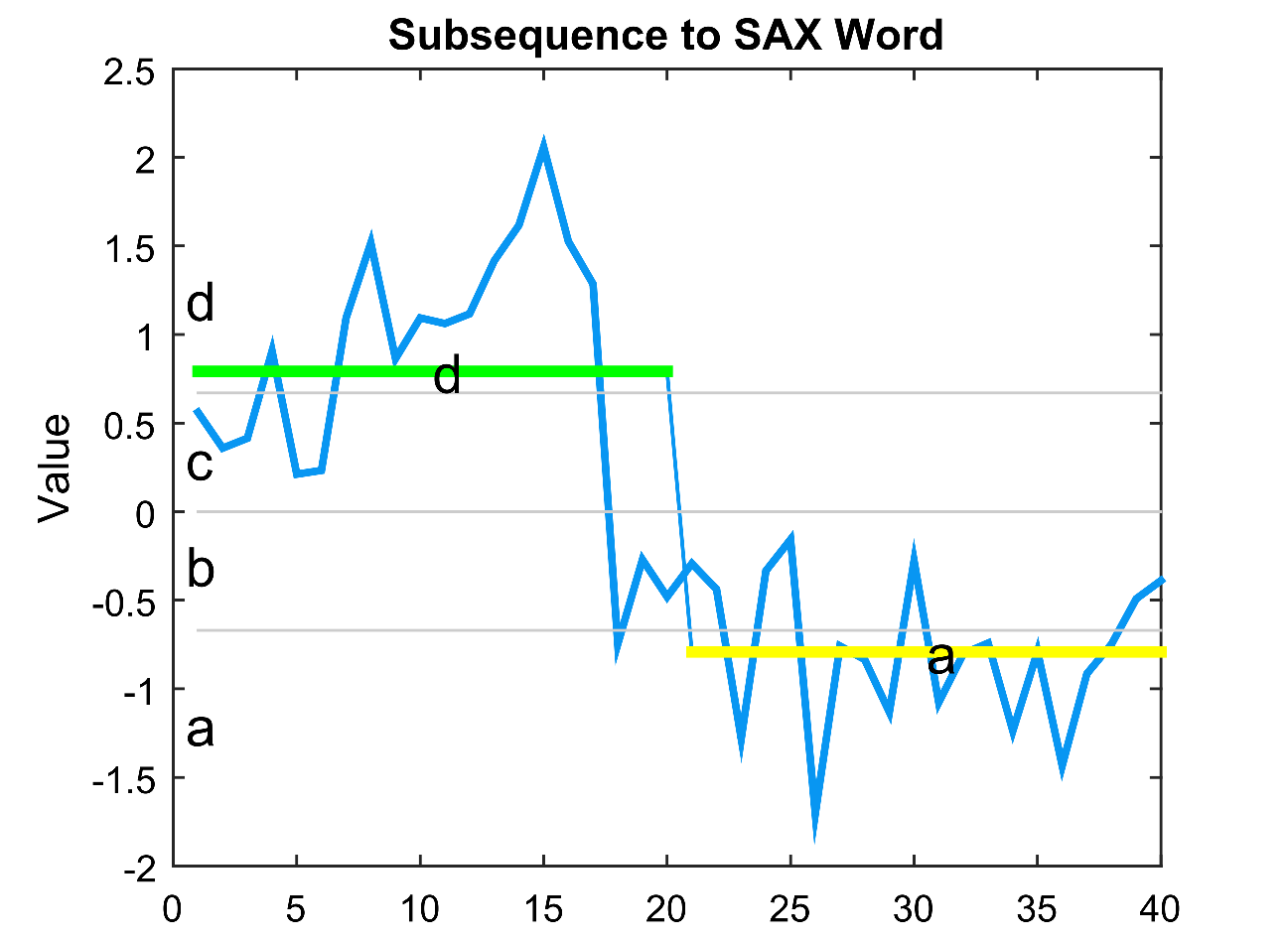}
  \caption{Transforming a subsequence to a symbolic pattern with Symbolic Aggregate approXimation (SAX)\cite{b4}\cite{b1}}
\end{figure}

The mean value for each segment is computed (the green line and yellow line in the figure for the two segments respectively). These mean values are mapped to symbols according to a set of break points (the gray lines in the figure). These break points divide the value space in equal-probable regions. In this example the alphabet size of SAX is 4 (with an alphabet of ‘a’, ‘b’, ‘c’ and ‘d’). The subsequence in the figure is transformed to the symbolic pattern “da”. The alphabet size $\gamma$ , number of segments (word length) $\omega$, and subsequence length l are supplied by the users.

\subsection{Symbolic Pattern Forest}

Symbolic Pattern Forest (SPF) \cite{b1}, which has linear time complexity. The approach checks if some randomly selected symbolic patterns exist in the time series to partition the data instances. This partition process is executed multiple times, and the partitions are combined by an ensemble process to generate the final partition. Figure 3 shows the framework structure of the proposed method. It was demonstrated that group structures in the data can emerge from the random partition process. Further analysis showed that the ensemble size needed to achieve  good results does not directly depend on the input data size, and thus the ensemble size was set to a proper fixed value for a specific data pattern.

\par

\begin{figure}[b]
  \centering
  \includegraphics[width=\linewidth]{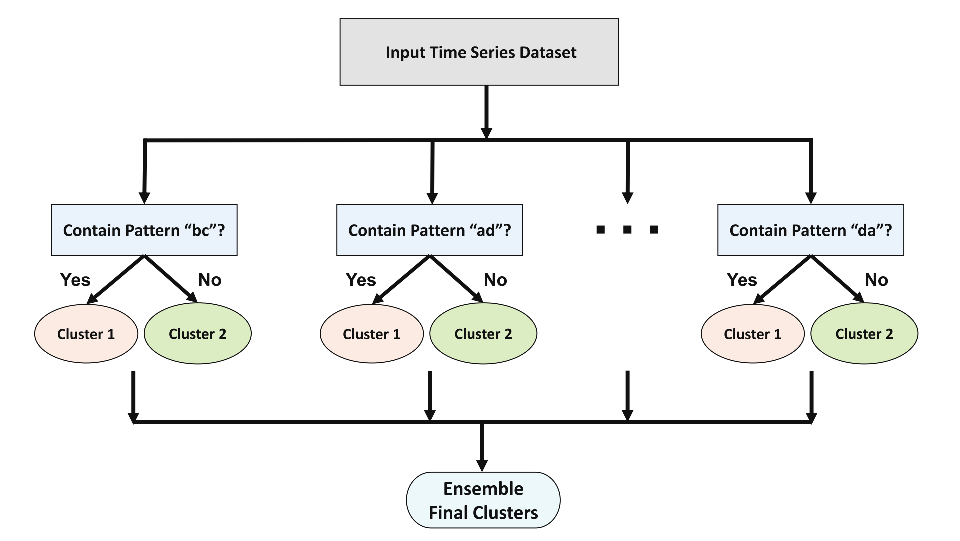}
  \caption{Framework of SPF, each branch in the figure is a tree and all the branches constitute a forest\cite{b1}}
\end{figure}

\begin{figure*}[t]
  \centering
  \includegraphics[width=\linewidth]{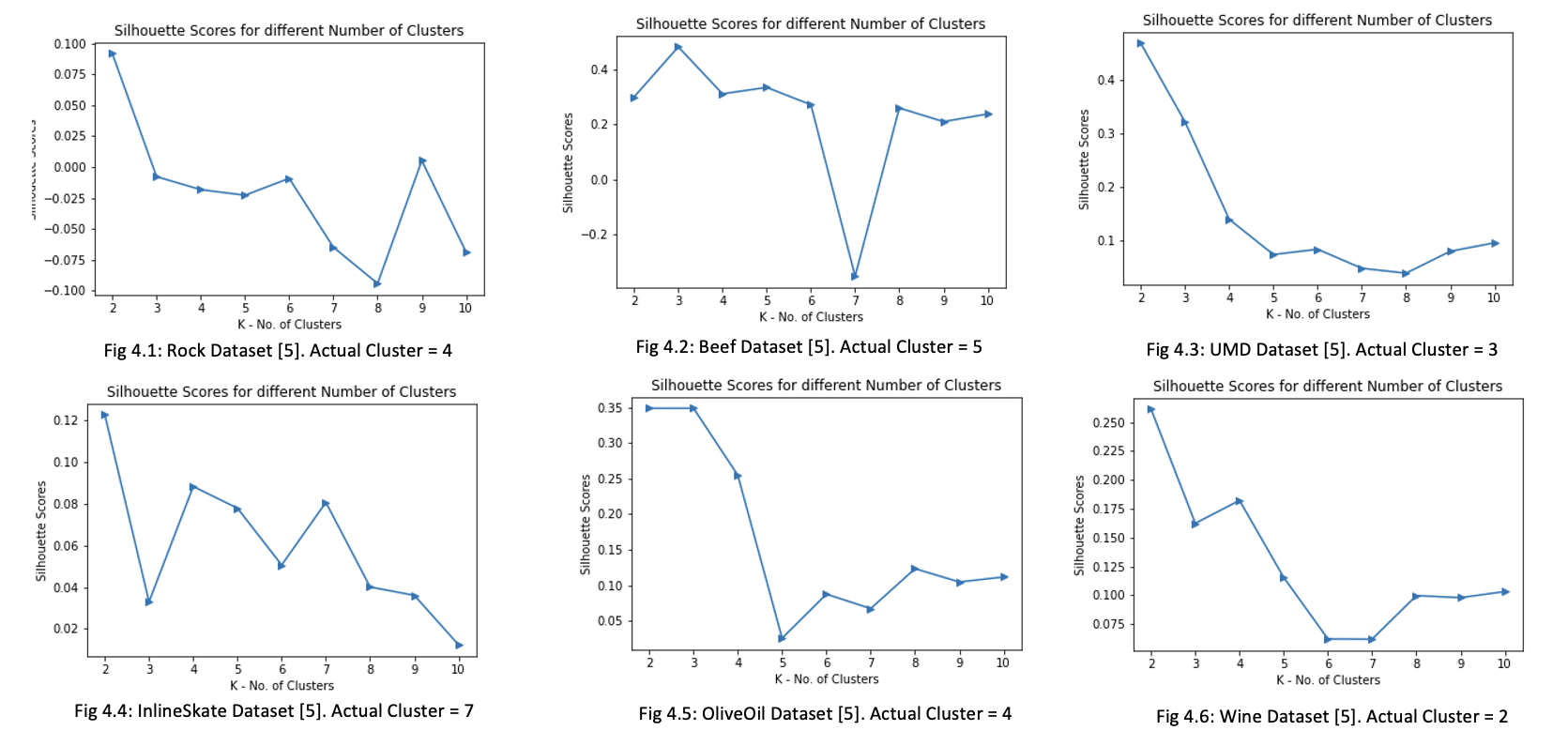}
  \caption{Predicting the optimal number of clusters based on Silhouette score on the raw datasets}
\end{figure*}

Further, the utilization of symbolic patterns makes the pattern space finite, and the symbolic patterns can be used to partition the data without using a distance measure. Checking the boolean indicating array to assign clusters in SPF is efficient as boolean operations are very fast. Boolean values are space-efficient which can take more advantage of the CPU cache to speed up the program.

\section{Our Methodologies}

\subsection{Experimenting with Raw Datasets}

Silhouette Coefficient was chosen as the clustering metric for our work. The equation to calculate the Silhouette score on any dataset is – 

$S(i) = ( b(i) – a(i) ) / max (b(i),a(i))$

where b(i) is the smallest average distance of point i to all points in any other cluster and a(i) is the average distance of i from all other points in its cluster. 

The value for Silhouette Coefficient ranges from -1 to 1.
\begin{itemize}
    \item 1: This means clusters are very well separated 	from each other.
	\item 0: This means the distances between the clusters are not that significant.
	\item -1: This means clusters are not assigned correctly, with lots of misclassifications.
\end{itemize}
	
\par

To the best of our knowledge, Silhouette coefficient was not used to determine the optimal number of clusters for any time series datasets before. And so, initially, we tested it on the raw time series and analyzed the results. We first used SPF \cite{b1} to predict the cluster labels for multiple number of clusters (K=2 to 10) and then we used the results with the raw data to determine the optimal K for the datasets. But as we can see in Figure 4, the predicted number of clusters are far off from the actual number of clusters. We concluded the fact, Silhouette on raw time series datasets is not an efficient clustering metric.

\subsection{Experimenting with Bag of Word Vectors}

A bag-of-words model is a way of extracting features from text for use in modeling, such as with machine learning algorithms. The approach is very simple and flexible, and can be used in a myriad of ways for extracting features from documents.The Bag of Words approach has been used in Natural Language Processing works for a very long time. Inspired by the success of text categorization \cite{b19} \cite{b20}, a bag-of-words representation became one of the most widely used methods for representing image content and has been successfully applied to object categorization. In this work, we used the Bag of Words concept and implemented on the SAX words that we obtained from the original time series subsequences.

\par

We used the SAX \cite{b4} algorithm to transform the time series subsequences to symbolic patterns (SAX Words), as briefly described in section \ref{SAX}. We used them to generate a vocabulary for all the unique words found in the dataset. The vectors are then generated for each time series, having the same length as the vocabulary size and each value in the vectors corresponds to the total number of times a word appeared in that specific time series. A few Bag of Word vectors generated from the Rock dataset\cite{b5} is shown in Table \ref{bow}.

\begin{figure*}[t]
  \centering
  \includegraphics[width=\linewidth]{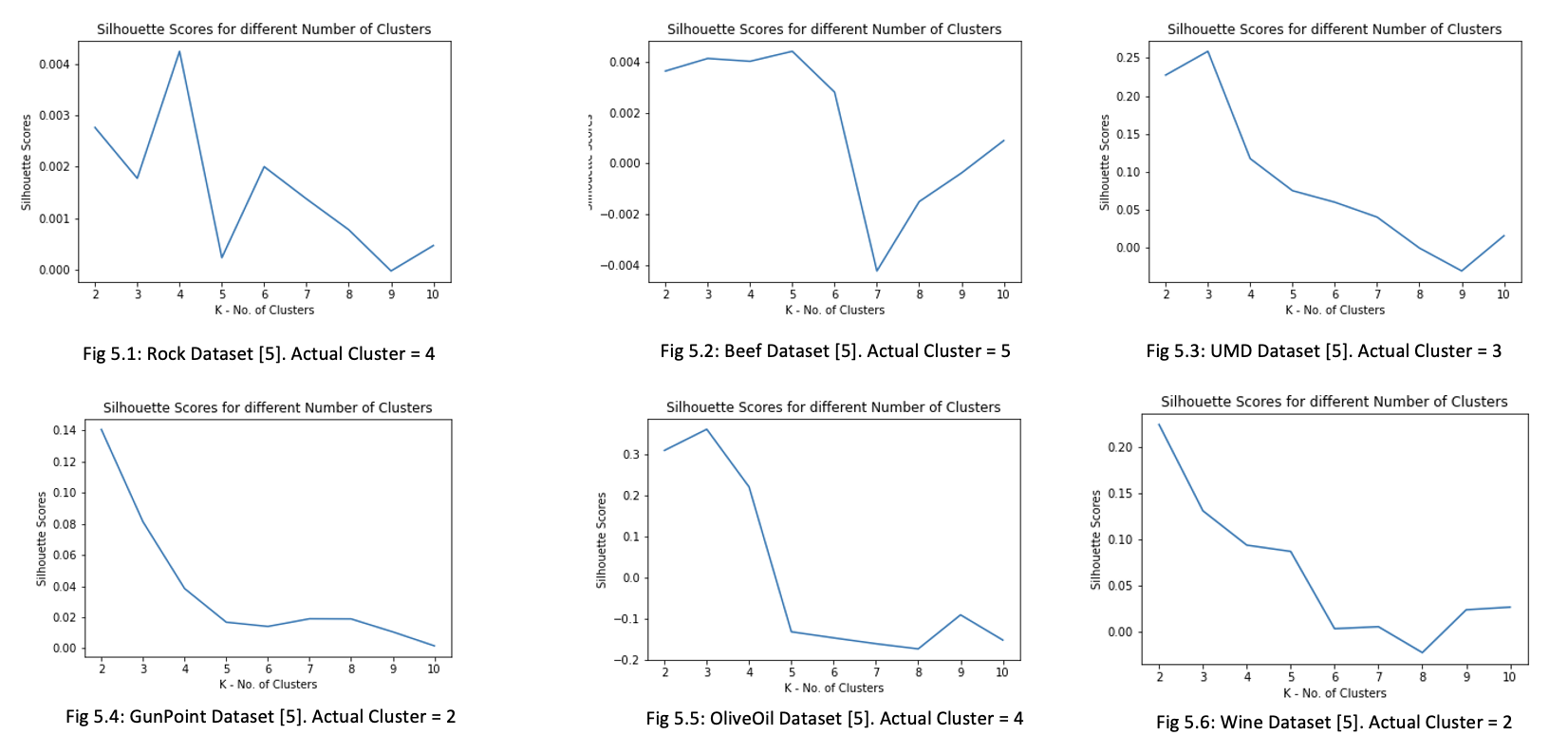}
  \caption{Predicting the optimal number of clusters based on Silhouette score on the BoW Vectors}
\end{figure*}

\begin{table}[b]
    \centering
    \begin{tabular}{cccccc} 
     \toprule
     & abbaa & caadb & aaabb & .. & ddddb\\ 
     \midrule
     \textbf{Bow Vector for TS1} & 0 & 1 & 1& .. & 3\\ 
     \textbf{Bow Vector for TS2} & 0 & 3 & 0& .. & 2\\ 
     \textbf{Bow Vector for TS3} & 0 & 2 & 0& .. & 0\\  
     \textbf{Bow Vector for TS4} & 2 & 0 & 0& .. & 0\\  
     \textbf{Bow Vector for TS5} & 1 & 0 & 0&  .. & 0\\  
     \textbf{..} & .. & .. & .. & .. & ..\\  
     \textbf{Bow Vector for TS60} & 3 & 0 &1 & .. & 0\\  
     \bottomrule
    \end{tabular}
    \caption{BoW vectors generated from Time Series}
    \label{bow}
\end{table}

Now, based on the cluster labels generated by SPF \cite{b1} for the different number of clusters, the Silhouette score was calculated on the Bag of Word vectors. The two SAX parameters - window size and alphabet size were not kept fixed for any of the datasets, since every dataset contains completely different types of time series with unique properties. Hence, within a range, all the combination of the two parameters were tested and the Silhouette score was calculated for every case. From all the results, the maximum value for Silhouette is then taken and the number of clusters where this maximum value was generated is then taken as the optimal value of K for the corresponding dataset. The results showed a significant improvement over the baseline.

\subsection{Experimenting with TF-IDF Vectors}

\par

Since the experiments on the Bag of Words vectors had promising results, we extended our work to TF-IDF vectors, which also is very popular technique in Natural Language Processing works. One issue with the Bag of Word vectors is that highly frequent words start to dominate, but they might not contain as much “informational content” as rarer but perhaps domain specific words. TF-IDF solves this issue by re-scaling the frequency of words by how often they appear. TF is a measure of the frequency of a word (w) in time series (ts).

\begin{center}
   \vspace{.2cm}
    $TF (w, ts) = \frac{\text{occurecnces of w in ts}}{\text{total number of w in ts}}$
    \vspace{.2cm} 
\end{center}

Whereas, IDF is the measure of the importance of a word.

\begin{center}
   \vspace{.2cm}
    $IDF (w, D) = ln \frac{\text{total number of ts in Dataset(D)}}{\text{total number ts containing w}}$
    \vspace{.2cm} 
\end{center}

The product of them is called TF-IDF and it serves two very important purposes -

\begin{itemize}
    \item gives more weightage to the word that is rare in the dataset.
    \item provides more importance to the word that is more frequent.
\end{itemize}

The scores are a weighting where not all words are equally as important or interesting. The scores have the effect of highlighting words that are distinct (contain useful information) in a given document. Thus the idf of a rare term is high, whereas the idf of a frequent term is likely to be low. A few tf-idf vectors generated from the Rock dataset\cite{b5} is shown in Table \ref{tfidf}.

Also, there are two more parameters that can be tuned to generate these vectors, called the minimum frequency and the maximum frequency. The words with the frequencies ranging between these two parameters are the ones that will appear in the vectors.

\begin{table}[b]
    \centering
    \begin{tabular}{cccccc} 
     \toprule
     & abbaa & caadb & aaabb & .. & ddddb\\ 
     \midrule
     \textbf{TF-IDF Vector for TS1} & 0.00 & 0.15 & 0.12& .. & 0.65\\ 
     \textbf{TF-IDF Vector for TS2} & 0.00 & 0.59 & 0.00& .. & 0.41\\ 
     \textbf{TF-IDF Vector for TS3} & 0.00 & 0.19 & 0.00& .. & 0.00\\  
     \textbf{TF-IDF Vector for TS4} & 0.57 & 0.00 & 0.00& .. & 0.00\\  
     \textbf{TF-IDF Vector for TS5} & 0.11 & 0.00 & 0.00&  .. & 0.00\\  
     \textbf{..} & .. & .. & .. & .. & ..\\  
     \textbf{TF-IDF Vector for TS60} & 0.49 & 0.00 &0.00 & .. & 0.00\\  
     \bottomrule
    \end{tabular}
    \caption{TF-IDF vectors generated from Time Series}
    \label{tfidf}
\end{table}

\begin{figure*}[t]
  \centering
  \includegraphics[width=\linewidth]{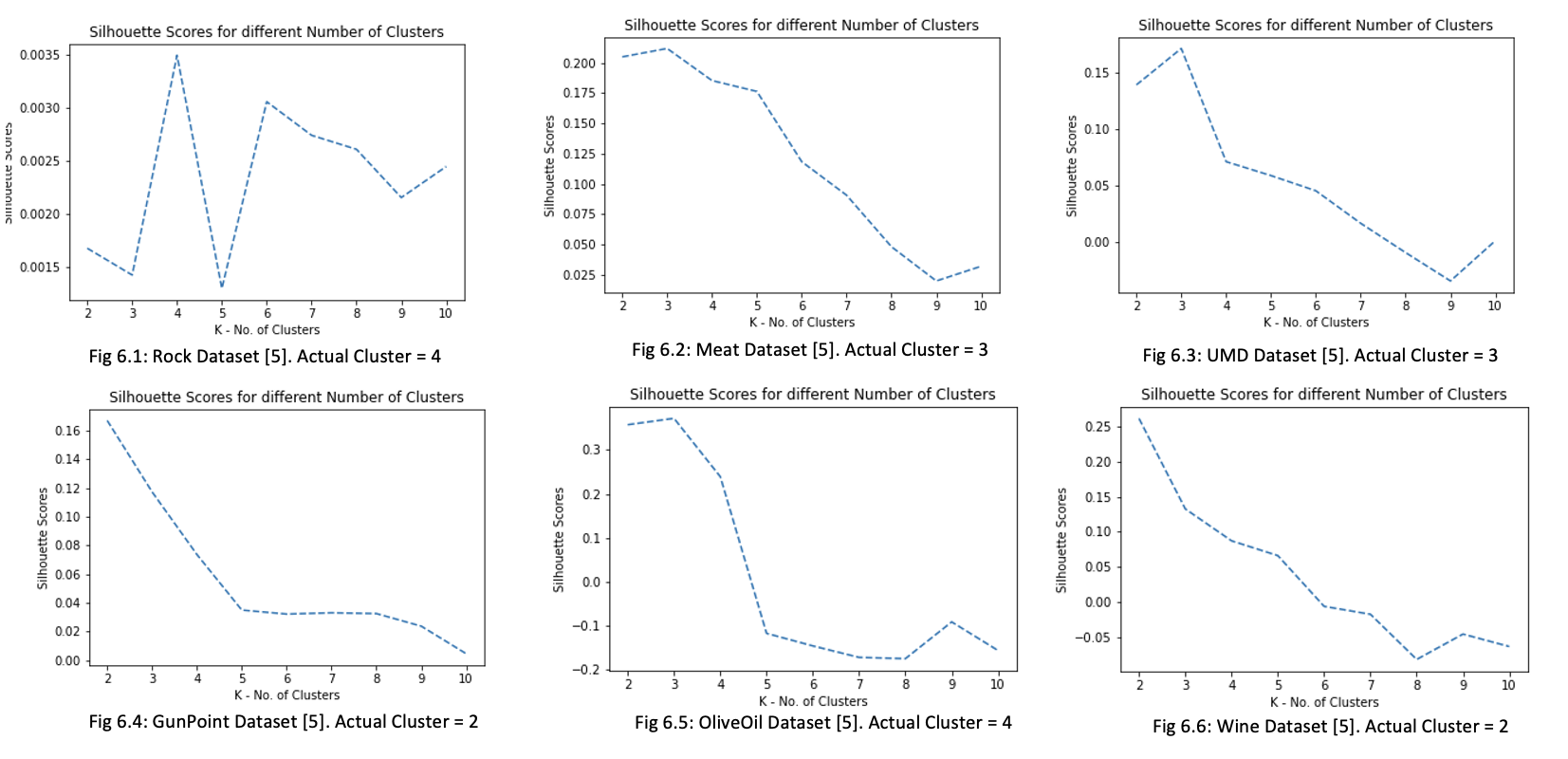}
  \caption{Predicting the optimal number of clusters based on Silhouette score on the TF-IDF Vectors}
\end{figure*}

Now, based on the cluster labels generated by SPF \cite{b1} for the different number of clusters, the Silhouette score was calculated on the TF-IDF vectors. The two SAX parameters - window size and alphabet size were kept as the same ones as the Bag of Word vectors for corresponding datasets. The two other parameters for tf-idf namely min frequency and max frequency was tested over a range. From all the results, the maximum value for Silhouette is then taken and the number of clusters where this maximum value was generated is then taken as the optimal value of K for the corresponding dataset. These results also showed a significant improvement over the baseline and remained almost consistent with the BoW vectors results.

\section{Performance Evaluation}

In our experiments, we predicted the optimal number of clusters in three different ways -

\begin{itemize}
    \item On raw dataset
    \item On Bag of Word Vectors
    \item On TF-IDF vectors
\end{itemize}

We determined the performances in these three cases based on how many times we got correct number of clusters, how many times the predicted results were close - meaning one more or less than the ground truth and how many times they were wrong.

\subsection{Baseline Results - On Raw Datasets}

Initially, the optimal number of clusters were predicted on raw datasets. Silhouette scores on the raw time series were quite consistent. It failed to predict the optimal number of clusters in most cases. We experimented on 30 datasets from the UCR archive \cite{b5} so far and the results presented in table \ref{raw} shows unconvincing results. Although a few of them were correct and some got close, based on the number of times they predicted correct, it can be concluded that Silhouette on raw dataset is not very dependable metric. 
\begin{table}[t]
    \centering
    \begin{tabular}{cccc} 
     \toprule
     Dataset & Actual Cluster & Predicted & Remarks\\ 
     \midrule
     \textbf{Beef} & 5 & 3 & Wrong\\ 
     \textbf{FaceFour} & 4 & 3 & Close\\ 
     \textbf{Fish} & 7 & 2 & Wrong\\ 
     \textbf{GunPoint} & 2 & 2 & \textbf{Correct}\\ 
     \textbf{Rock} & 4 & 2 & Wrong\\ 
     \textbf{HouseTwenty} & 2 & 4 & Wrong\\ 
     \textbf{EthanolLevel} & 4 & 2 & Wrong\\ 
     \textbf{Wine} & 2 & 2 & \textbf{Correct}\\ 
     \textbf{Wafer} & 2 & 2 & \textbf{Correct}\\ 
     \textbf{SyntheticControl} & 6 & 3 & Wrong\\ 
     \textbf{InlineSkate} & 7 & 2 & Wrong\\ 
     \textbf{InsectEPGRegularTrain} & 3 & 2 & Close\\ 
     \textbf{GunPointAgeSpan} & 2 & 8 & Wrong\\ 
     \textbf{Haptics} & 5 & 2 & Wrong\\ 
     \textbf{UMD} & 3 & 2 & Close\\ 
     \textbf{Symbols} & 6 & 3 & Wrong\\ 
     \textbf{OliveOil} & 4 & 2 & Wrong\\ 
     \textbf{HandOutlines} & 2 & 2 & \textbf{Correct}\\ 
     \textbf{Meat} & 3 & 2 & Close\\ 
     \textbf{ECG200} & 2 & 3 & Close\\ 
     \textbf{WormsTwoClass} & 2 & 4 & Wrong\\ 
     \textbf{Worms} & 5 & 4 & Close\\ 
     \textbf{Plane} & 7 & 6 & Close\\ 
     \textbf{Strawberry} & 2 & 2 & \textbf{Correct}\\ 
     \textbf{Trace} & 4 & 2 & Wrong\\ 
     \textbf{Lightning7} & 7 & 5 & Wrong\\ 
     \textbf{MoteStrain} & 2 & 4 & Wrong\\ 
     \textbf{ChinaTown} & 2 & 2 & \textbf{Correct}\\ 
     \textbf{TwoPatterns} & 4 & 2 & Wrong\\ 
     \textbf{TwoLeadECG} & 2 & 4 & Wrong\\ 
     \bottomrule
    \end{tabular}
    \caption{Silhouette on Raw Time Series}
    \label{raw}
\end{table}

\begin{table*}[t]
    \centering
    \begin{tabular}{cccccc} 
     \toprule
     Dataset & SAX Window Size & SAX Alphabet Size & Actual Clusters & Predicted Clusters & Remarks\\ 
     \midrule
     \textbf{Beef} & 5 & 8 & 5 & 5 & \textbf{Correct}\\ 
     \textbf{FaceFour} & 40 & 8 & 4 & 3 & Close\\ 
     \textbf{Fish} & 5 & 20 & 7 & 8 & Close \\ 
     \textbf{GunPoint} & 5 & 9 & 2 & 2 & \textbf{Correct}\\ 
     \textbf{Rock} & 8 & 5 & 4 & 4 & \textbf{Correct}\\ 
     \textbf{HouseTwenty} & 50 & 8 & 2 & 2 & \textbf{Correct}\\ 
     \textbf{EthanolLevel} & 350 & 10 & 4 & 3 & Close\\ 
     \textbf{Wine} & 3 & 4 & 2 & 2 & \textbf{Correct}\\ 
     \textbf{Wafer} & 20 & 5 & 2 & 2 & \textbf{Correct}\\ 
     \textbf{SyntheticControl} & 3 & 6 & 6 & 6 & \textbf{Correct}\\ 
     \textbf{InlineSkate} & 100 & 8 & 7 & 2 & Wrong\\ 
     \textbf{InsectEPGRegularTrain} & 100 & 3 & 3 & 2 & Close\\ 
     \textbf{GunPointAgeSpan} & 10 & 8 & 2 & 2 & \textbf{Correct}\\ 
     \textbf{Haptics} & 20 & 10 & 5 & 2 & Wrong\\ 
     \textbf{UMD} & 20 & 8 & 3 & 3 & \textbf{Correct}\\ 
     \textbf{Symbols} & 30 & 10 & 6 & 3 & Wrong\\ 
     \textbf{OliveOil} & 100 & 4 & 4 & 3 & Close\\ 
     \textbf{HandOutlines} & 100 & 4 & 2 & 2 & \textbf{Correct}\\ 
     \textbf{Meat} & 20 & 4 & 3 & 2 & Close\\ 
     \textbf{ECG200} & 100 & 4 & 2 & 2 & \textbf{Correct}\\ 
     \textbf{WormsTwoClass} & 200 & 10 & 2 & 2 & \textbf{Correct}\\ 
     \textbf{Worms} & 200 & 10 & 5 & 2 & Wrong\\ 
     \textbf{Plane} & 10 & 4 & 7 & 7 & \textbf{Correct}\\ 
     \textbf{Strawberry} & 50 & 4 & 2 & 2 & \textbf{Correct}\\ 
     \textbf{Trace} & 50 & 4 & 4 & 4 & \textbf{Correct}\\ 
     \textbf{Lightning7} & 100 & 8 & 7 & 4 & Wrong\\ 
     \textbf{MoteStrain} & 5 & 4 & 2 & 2 & \textbf{Correct}\\ 
     \textbf{ChinaTown} & 12 & 3 & 2 & 2 & \textbf{Correct}\\ 
     \textbf{TwoPatterns} & 50 & 4 & 4 & 2 & Wrong\\ 
     \textbf{TwoLeadECG} & 40 & 8 & 2 & 2 & \textbf{Correct}\\ 
     \bottomrule
    \end{tabular}
    \caption{Silhouette on BoW Vectors}
    \label{bowvectors}
\end{table*}

\subsection{BoW Vector Results}

Next, as mentioned in section IV-B, we experimented with the Bag of Word vectors. The vectors were generated from the SAX words of the time series. The two parameters of the SAX algorithm \cite{b2} were not kept fixed for any datasets. For every datasets a range of the two parameters were tested and the combination that resulted in the maximum silhouette score determined the optimal number of clusters for that specific dataset. We presented the SAX parameters with their corresponding datasets in table IV. In the table, we also showed the results we got from different datasets. Compared to the baseline results, we observed a significant improvement in terms of predicting the right number of clusters. The results were consistent throughout the datasets and the number of times wrong prediction generated was fairly low compared to before.

\subsection{TF-IDF Vector Results}

Finally, as mentioned in section IV-C, we generated the TF-IDF vectors from the SAX words. The two SAX parameters were kept the same which resulted the maximum Silhouette score for the Bag of Words vectors. But case of the TF-IDF vectors we worked with two other parameters - minimum frequency and maximum frequency. Any word that had less frequency than the minimum frequency was not taken into account while generating the vectors and also any word that had a higher frequency than the maximum were omitted. The results on the 30 datasets that we worked with so far were consistent with the results from Bag of Word vectors with a slight better performance. This also meant a significant improvement over the baseline. The results along with the parameters chosen for the specific datasets are presented in table V.

\subsection{Comparison}

When we compared the results obtained from all three scenarios, the baseline results were fairly poor, having predicted the correct number of clusters only 20\% of the time and getting it wrong 58\% of the time. Experiments with the Bag of Word vectors resulted with correct results 60\% of the time and got it wrong only 20\% of the times. The results from the tf-idf vectors were almost consistent with the Bag of Word vectors, having a slightly better accuracy. All the results are shown in Figure 7. Overall, from all the observations, it can be concluded that there is a significant correlation between the number of clusters for a time series datasets and their corresponding Bag of Word and TF-IDF vectors.

\begin{table*}[t]
    \centering
    \begin{tabular}{cccccccc} 
     \toprule
     Dataset & SAX Window Size & SAX Alphabet Size & Min Freq & Max Freq & Actual Clusters & Predicted Clusters & Remarks\\ 
     \midrule
     \textbf{Beef} & 5 & 8 & 0.01 & 0.9 & 5 & 5 & \textbf{Correct}\\ 
     \textbf{FaceFour} & 40 & 8 & 0.1 & 0.9 & 4 & 3 & Close\\ 
     \textbf{Fish} & 5 & 20 & 0.001 & 0.01 &7 & 8 & Close \\ 
     \textbf{GunPoint} & 5 & 9 & 0.01 & 0.9 &2 & 2 & \textbf{Correct}\\ 
     \textbf{Rock} & 8 & 5 & 0.01 & 0.9 &4 & 4 & \textbf{Correct}\\ 
     \textbf{HouseTwenty} & 50 & 8 & 0.01 & 0.9 &2 & 2 & \textbf{Correct}\\ 
     \textbf{EthanolLevel} & 350 & 10 & 0.001 & 0.99 &4 & 3 & Close\\ 
     \textbf{Wine} & 3 & 4 & 0.1 & 0.9 &2 & 2 & \textbf{Correct}\\ 
     \textbf{Wafer} & 20 & 5 & 0.001 & 0.99 &2 & 2 & \textbf{Correct}\\ 
     \textbf{SyntheticControl} & 3 & 6 & 0.1 & 0.9 &6 & 6 & \textbf{Correct}\\ 
     \textbf{InlineSkate} & 100 & 8 & 0.15 & 0.8 &7 & 2 & Wrong\\ 
     \textbf{InsectEPGRegularTrain} & 100 & 3 & 0.25 & 0.9 &3 & 2 & Close\\ 
     \textbf{GunPointAgeSpan} & 10 & 8 & 0.001 & 0.9 &2 & 2 & \textbf{Correct}\\ 
     \textbf{Haptics} & 20 & 10 & 0.2 & 0.95 &5 & 2 & Wrong\\ 
     \textbf{UMD} & 20 & 8 & 0.001 & 0.99 &3 & 3 & \textbf{Correct}\\ 
     \textbf{Symbols} & 30 & 10 & 0.001 & 0.99 &6 & 3 & Wrong\\ 
     \textbf{OliveOil} & 100 & 4 & 0.1 & 0.9 &4 & 3 & Close\\ 
     \textbf{HandOutlines} & 100 & 4 & 0.1 & 0.9 &2 & 2 & \textbf{Correct}\\ 
     \textbf{Meat} & 20 & 4 & 0.1 & 0.9 &3 & 2 & Close\\ 
     \textbf{ECG200} & 100 & 4 & 0.1 & 0.9 &2 & 2 & \textbf{Correct}\\ 
     \textbf{WormsTwoClass} & 200 & 10 &0.1 & 0.9 & 2 & 2 & \textbf{Correct}\\ 
     \textbf{Worms} & 200 & 10 & 0.1 & 0.9 &5 & 2 & Wrong\\ 
     \textbf{Plane} & 10 & 4 & 0.01 & 0.99 &7 & 7 & \textbf{Correct}\\ 
     \textbf{Strawberry} & 50 & 4 &0.01 & 0.99 & 2 & 2 & \textbf{Correct}\\ 
     \textbf{Trace} & 50 & 4 & 0.01 & 0.99 &4 & 4 & \textbf{Correct}\\ 
     \textbf{Lightning7} & 100 & 80 & 0.01 & 0.99 & 7 & 4 & Wrong\\ 
     \textbf{MoteStrain} & 5 & 4 &0.01 & 0.99 & 2 & 2 & \textbf{Correct}\\ 
     \textbf{ChinaTown} & 12 & 3 & 0.01 & 0.99 &2 & 2 & \textbf{Correct}\\ 
     \textbf{TwoPatterns} & 50 & 4 & 0.01 & 0.99 &4 & 2 & Wrong\\ 
     \textbf{TwoLeadECG} & 40 & 8 &0.01 & 0.99&2 & 2 & \textbf{Correct}\\ 
     \bottomrule
    \end{tabular}
    \caption{Silhouette on TF-IDF Vectors}
    \label{tfidf-vectors}
\end{table*}

\begin{figure}[t]
  \centering
  \includegraphics[width=\linewidth]{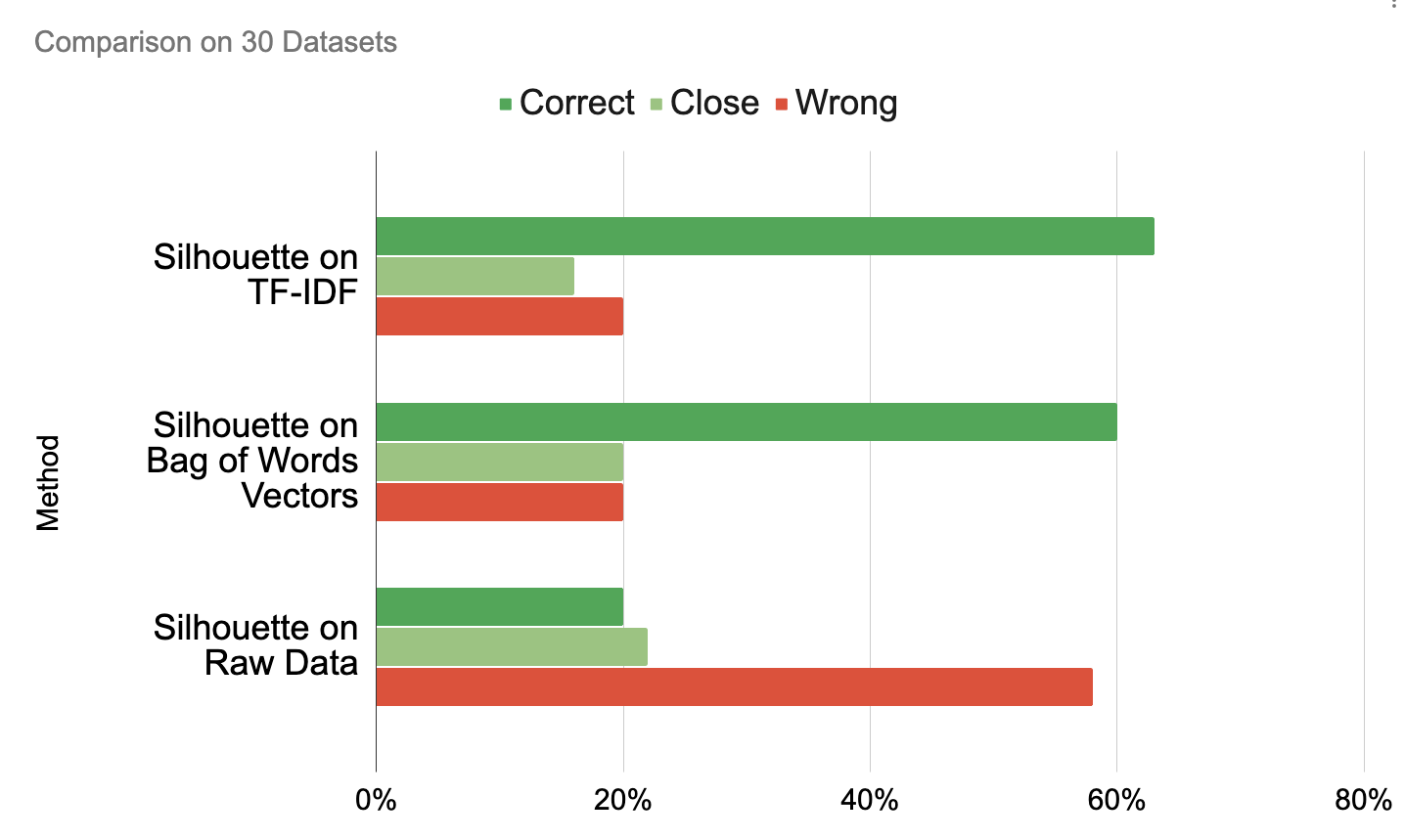}
  \caption{Comparing Results}
\end{figure}

\section{Conclusion and Future Works}

In this research endeavor, we have meticulously expanded upon the Symbolic Pattern Forest algorithm, as delineated in \cite{b1}, with the primary objective of predicting the optimal number of clusters specifically for time series datasets. This is particularly crucial in scenarios where the ground truth pertaining to the actual number of clusters remains elusive or unknown. Our extensive study and analysis have demonstrated that, despite the prevalent utilization of the Silhouette Score as a pivotal clustering metric to ascertain the qualitative aspects of cluster numbers, it unfortunately does not serve as an efficacious metric when applied directly to raw time series datasets.

However, a significant enhancement in results was observed when the silhouette was applied to our refined methodology. In our approach, we employed the Symbolic Aggregate approXimation (SAX) words to meticulously generate Bag of Words vectors and subsequently, TF-IDF vectors. This innovative approach yielded results that were markedly superior compared to the baseline methodologies, showcasing the efficacy of our proposed enhancements in clustering methodologies. It is our intention to further this experimentation on an extensive array of datasets, specifically all the 128 datasets available from the UCR archive as referenced in \cite{b5}. By doing so, we aim to further corroborate and solidify the preliminary results that we have garnered thus far.

Moreover, we are committed to refining our methodologies and approaches to enhance the precision in predicting the optimal number of clusters. This is pivotal for ensuring the reliability and robustness of clustering in time series datasets, which is crucial for extracting meaningful insights from the data. By achieving a higher level of accuracy, we can ensure that the derived clusters are more representative of the inherent structures within the datasets, thereby leading to more reliable and insightful conclusions. Our ongoing work is poised to contribute significantly to the field of time series analysis by providing enhanced methodologies for optimal cluster number prediction, which is instrumental in unveiling the intricate patterns and structures embedded within time series data.

\end{document}